\documentclass[runningheads]{llncs}
\usepackage{graphicx}
\usepackage{bbm}
\usepackage{multirow}
\usepackage{hyperref}
\usepackage{xcolor}
\usepackage{booktabs}

\begin{document}
\title{FBA-Net: Foreground and Background Aware Contrastive Learning for Semi-Supervised Atrium Segmentation}
\titlerunning{FBA-Net}
%
\author{Yunsung Chung\inst{1} \and Chanho Lim\inst{2} \and Chao Huang\inst{2} \and Nassir Marrouche\inst{2} \and Jihun Hamm\inst{1}}
\authorrunning{C. Yunsung et al.}
%
\institute{Department of Computer Science, Tulane University, New Orleans, USA \and
School of Medicine, Tulane University, New Orleans, USA\\
\email{\{ychung3, clim, chuang16, nmarrouche, jhamm3\}@tulane.edu}}
%
\maketitle              
\begin{abstract}
Medical image segmentation of gadolinium enhancement magnetic resonance imaging (GE MRI) is an important task in clinical applications. However, manual annotation is time-consuming and requires specialized expertise. Semi-supervised segmentation methods that leverage both labeled and unlabeled data have shown promise, with contrastive learning emerging as a particularly effective approach. In this paper, we propose a contrastive learning strategy of foreground and background representations for semi-supervised 3D medical image segmentation (FBA-Net). Specifically, we leverage the contrastive loss to learn representations of both the foreground and background regions in the images. By training the network to distinguish between foreground-background pairs, we aim to learn a representation that can effectively capture the anatomical structures of interest. Experiments on three medical segmentation datasets demonstrate state-of-the-art performance. Notably, our method achieves a Dice score of 91.31\% with only 20\% labeled data, which is remarkably close to the 91.62\% score of the fully supervised method that uses 100\% labeled data on the left atrium dataset. Our framework has the potential to advance the field of semi-supervised 3D medical image segmentation and enable more efficient and accurate analysis of medical images with a limited amount of annotated labels. Our code is available at \url{https://github.com/cys1102/FBA-Net}.
\keywords{Semi-supervised learning  \and Contrastive learning \and Cardiac Image segmentation.}
\end{abstract}
\section{Introduction}
Medical image segmentation of the left atrium (LA) plays a crucial role in many clinical applications, including diagnosis, treatment planning, and disease monitoring. In recent years, deep learning approaches \cite{chen2021jas,li2022atrialjsqnet,yang2020simultaneous,liu2022ugformer} have shown promising results in medical image segmentation tasks but require a large amount of manually annotated data for training. In addition, manual annotation of medical images is a challenging task that requires expert knowledge and specialized tools, leading to time-consuming and laborious procedures. The need for utilizing unannotated data has motivated researchers to develop semi-supervised learning techniques that can leverage both labeled and unlabeled data to improve accuracy.\\
As a solution, contrastive learning \cite{bachman2019learning,chen2020simple,he2020momentum,chen2020improved,xie2022contrastive} has emerged as a promising approach for downstream tasks with unlabeled data to obtain a better initialization in various computer vision tasks, including medical image processing \cite{azizi2021big,cho2023chess,zhou2021preservational,azizi2022robust,kiyasseh2021segmentation,bai2019self,tang2022self}. This technique leverages unlabeled data to learn meaningful representations that capture the underlying structure of the data. These representations can be used to improve the performance of supervised learning algorithms on labeled data. The contrastive learning strategy has also been employed for semi-supervised learning in medical image segmentation \cite{you2022momentum,zhao2023rcps}. \\
Semi-supervised learning with contrastive learning has gained popularity in recent years for its ability to reduce the burden of annotation. However, we believe that two significant issues have been neglected in existing investigations. Firstly, many of the existing methods focus on relationships between voxels, which require considerable computational resources and depend heavily on augmentation to generate positive pairs. Secondly, most contrastive learning studies disregard the specific characteristics of segmentation tasks when extracting representations. We suggest that tailored representations to meet the requirements of segmentation tasks could enhance performance with minimum additional computational costs. \\
To address these issues, we propose a semi-supervised learning approach that focuses on discriminating the foreground and background representations (FBA-Net) by contrastive learning. Our approach trains the model to distinguish between the foreground-background regions of target objects by optimizing contrastive loss. This enables the model to identify important foreground features while ignoring the less relevant background features, leading to better performance in segmenting target objects and extracting precise boundaries between them. By utilizing semi-supervised techniques, this approach offers a potential solution for reducing the dependence on labeled data while improving the accuracy of medical image analysis. \\
In this paper, we make the following contributions: (1) We propose a novel contrasting strategy of foreground and background representations specialized for medical image segmentation and leverage unannotated labels to alleviate the burden of annotation. (2) We introduce a contrastive module that allows the network to distinguish between foreground and background regions of target objects, reducing the reliance on consistency loss. The module is designed to be easily integrated into any network. (3) We evaluate the proposed method on three public datasets and observe that it outperforms existing state-of-the-art methods. Notably, our proposed method, when trained on just 20\% of labeled data, shows a minimal dice score difference of only 0.31\% compared to fully supervised learning which is trained on 100\% of labeled data on the LA dataset.
\subsubsection{Related Work.} In the field of contrastive learning, Bachman et al. \cite{bachman2019learning} introduced an autoregressive model to generate a context for multi-views of the same data to train a network by predicting the context of one view given the context of another view. Chen et al. \cite{chen2020simple} used data augmentation to produce different views from the same image to learn representations that are invariant to transformations. He et al. \cite{he2020momentum} introduced a momentum-based update rule to generate a dynamic dictionary of visual representations. \\
Semi-supervised learning approaches have been applied to medical image segmentation. Li et al. \cite{li2020shape} incorporated a shape prior into the network using the signed distance map to encode the shape information. Luo et al. \cite{luo2021semi} introduced a dual-task consistency method to enforce consistency between segmentation masks and an auxiliary task. Wu et al. \cite{wu2022mutual} proposed mutual consistency learning approach from multiple different decoders. You et al. \cite{you2022momentum} introduced a semi-supervised learning framework for volumetric medical image segmentation using momentum contrastive voxel-wise representation learning. Zhao et al. \cite{zhao2023rcps} proposed a voxel-level contrastive learning framework with intensity augmentations. \\ 
While contrastive learning and semi-supervised learning have shown promising results in medical image segmentation, FBA-Net differs from existing methods in two ways. Firstly, unlike recent works that generate positive pairs via augmentations of identical input and compute the contrastive loss between voxels. Instead, we employ a contrastive module to extract representations of target entities. This not only simplifies the training process by reducing computational resource demands but also lessens the dependence on augmentations techniques. Secondly, while previous contrastive learning methods have used a variation of InfoNCE loss \cite{oord2018representation}, our method adopts loss functions specially designed to differentiate between foreground and background representations in medical image segmentation.
\section{Method}
\subsection{Architecture}
FBA-Net is comprised of two major components: a pseudo-label generation module and a contrastive learning module. Given a dataset represented as $(X, Y)$, we have images $x\in X$ and their corresponding labels $y\in Y$. $X$ comprises $N$ labeled and $M$ unlabeled slices ($N\ll M$). From input images $x_i\in X$, the network extracts the foreground regions of target objects, denoted as $M_i\in \mathbbm{R}^{H\times W\times C}$. By creating $(1-M_i)\in \mathbbm{R}^{H\times W\times C}$, we can generate corresponding background regions. An encoder $h(\cdot)$ with a projection head is used to map $z_i^f$ and $z_i^b$ to process the foreground and background representations further, respectively. The projection head is responsible for projecting foreground and background maps onto a latent space where contrastive loss can be implemented.
\begin{figure}[t]
\includegraphics[width=\textwidth]{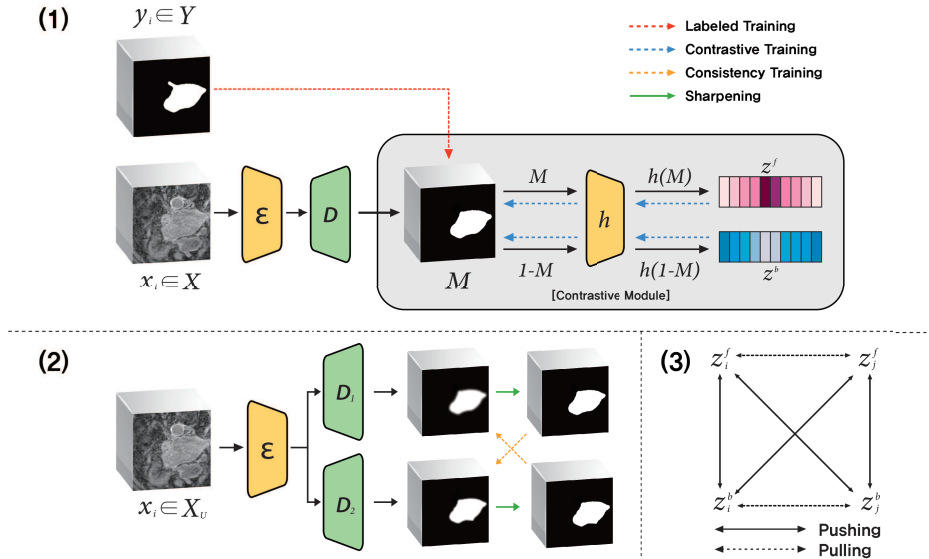}
\caption{A overview of FBA-Net includes the following aspects. (A) Our proposed contrastive training strategy: given an image $x\in X$, we can obtain the background from $1-M$, assuming $M$ is the foreground region. The encoder $h(\cdot)$ creates foreground and background representations, denoted by $z^f$ and $z^b$, respectively. (B) Mutual consistency training: two different decoders $D_1, D_2$ generate pseudo labels for each other for unlabeled images, $x_i \in X_U$. (3) Foreground and background representations, $z^f_i$ and $z^b_i$, respectively, are created as positive and negative contrastive pairs. The positive pairs are pulled closer, while the negative pairs are pushed apart.} \label{fig1}
\end{figure}
\subsection{Contrastive Learning}
FBA-Net employs contrastive learning, which learns representations of data by contrasting foreground and background representations. This approach can aid in identifying the precise boundaries of targets by maximizing the gap between foreground-background relationships. Additionally, contrastive learning can help alleviate the need for a large number of pixel-wise labels, which is a significant challenge in image segmentation.\\
Inspired by Xie et al. \cite{xie2022contrastive}, we introduce two distinct losses for positive and negative pairs. However, instead of channel-wise representations, we extract spatial-wise representations of foreground-background. As noted by \cite{chen2020simple,chen2020improved}, the network learns faster using contrastive loss with large batch size, which needs high computational costs. To achieve maximum effectiveness with smaller batches, we use ranking weights. We first compute the similarities between representations using the following equation:
\begin{equation}
s_{ij}=sim(z_i,z_j)
\end{equation}
where $sim$ indicates the cosine similarity function. Given the set of similarities $S_{ij}=\{s_{11},s_{12},...,s_{ij}\}$, the ranking weights are calculated as
\begin{equation}
w_{ij}=\exp(-\alpha \cdot rank(sim(z_i, z_j))
\end{equation}
where $\alpha$ is a hyperparameter that controls the smoothness of the exponential function. We empirically set $\alpha$ as 0.25. $rank$ denotes the rank function within $S_{i,j}$ and the weight ranges from 0 to 1.\\
The positive pairs are responsible for maximizing the similarity between representations. A foreground object in one image should be located closer to the foreground representation of another image in the semantic space. This principle similarly applies to background-background associations as shown in Fig~\ref{fig1}. Given $n$ input images, the contrastive module computes $n$ foreground and background representations, denoted as $z^f_n$ and $z^b_n$, respectively. Positive pairs are formed between the same foreground or background representations, excluding oneself. We define positive losses for foreground-foreground and background-background pairs as follows
\begin{equation}
\mathcal{L}_{pos}^{f}=-\frac{1}{n(n-1)}\sum^n_{i=1}\sum^n_{j=1}\mathbbm{1}_{\small[i\not=j\small]}\log(w^f_{ij}\cdot sim(z_i^f, z_j^f))
\end{equation}
\begin{equation}
\mathcal{L}_{pos}^{b}=-\frac{1}{n(n-1)}\sum^n_{i=1}\sum^n_{j=1}\mathbbm{1}_{\small[i\not=j\small]}\log(w^b_{ij}\cdot sim(z_i^b, z_j^b))
\end{equation}
where $\mathbbm{1}_{\small[i\not=j\small]}\in\{0,1\}$ represents an indicator function that outputs $1$ if $i\not=j$. The positive loss combines each loss of two positive foreground and background pairs. 
\begin{equation}
\mathcal{L}_{pos}=\mathcal{L}_{pos}^f+\mathcal{L}_{pos}^b
\end{equation}
The foreground and background representations have different meanings and play distinct roles in segmentation. To enlarge the difference, we use negative pair loss as part of our training process. The negative pair loss encourages the model to distinguish between foreground and background objects, allowing for more precise and accurate segmentation. The negative pair loss is defined as
\begin{equation}
\mathcal{L}_{neg}=-\frac{1}{n^2}\sum^n_{i=1}\sum^n_{j=1}\log(w^{f,b}_{ij}\cdot (1-sim(z_i^f, z_j^b)))
\end{equation}
The contrastive loss and the overall training loss are provided below:
\begin{equation}
\mathcal{L}_{contra}=\mathcal{L}_{pos}+\mathcal{L}_{neg},
\end{equation}
\begin{equation}
\mathcal{L}=\mathcal{L}_{dice}+\mathcal{L}_{contra}+\mathcal{L}_{consist},
\end{equation}
where $\mathcal{L}_{dice}$ and $\mathcal{L}_{consist}$ represent Dice loss for labeled training and MSE loss for mutual training.
\begin{table}[t]
    \centering
    \caption{Comparisons of semi-supervised segmentation models on LA, Pancreas-CT, and ACDC datasets. * indicates our re-implementation methods.}
    \label{tab:tb1}
    \begin{tabular}{@{}lcccccccc@{}}
        \toprule
        \multirow{2}{*}{Method} & \multirow{2}{*}{Labeled data} & \multicolumn{2}{c}{\textbf{LA}} & \multicolumn{2}{c}{\textbf{Pancreas-CT}} & \multicolumn{2}{c}{\textbf{ACDC}} \\
        \cmidrule(lr){3-4} \cmidrule(lr){5-6} \cmidrule(lr){7-8}
        &  & Dice $\uparrow$ & ASD$\downarrow$  & Dice $\uparrow$ & ASD$\downarrow$  & Dice $\uparrow$ & ASD$\downarrow$ \\
        \midrule
        Supervised          & 100\%                 & 91.62             & 1.64          & 82.60             & 1.33          & 91.65             & 0.56\\ 
        \midrule
        SSASSNet \cite{li2020shape}     & \multirow{7}{*}{10\%} & 85.81             & 4.04          & 68.97             & \textbf{1.96} & 84.14             & 1.40\\
        DTC \cite{luo2021semi}          &                       & 87.91             & 2.92          & 66.58             & 4.16          & 82.71             & 2.99\\
        CVRL* \cite{you2022momentum}    &                       & 88.06             & 3.11          & 69.03             & 3.95          & 86.66             & 3.27 \\
        MC-NET \cite{wu2021semi}        &                       & 87.92             & 2.64          & 69.06             & 2.28          & 86.34             & 2.08\\ 
        MC-NET+ \cite{wu2022mutual}     &                       & 88.39             & 1.99          & 70.00             & 3.87          & 87.10             & 2.00\\ 
        RCPS* \cite{zhao2023rcps}       &                       & \textbf{89.24}    & 2.12          & 71.24             & 3.71          & 88.09             & 1.96\\
        \textbf{FBA-Net}                &                       & 88.69             & \textbf{1.92} & \textbf{71.35}    & 3.00          & \textbf{88.45}    & \textbf{0.71}\\
        \midrule
        SSASSNet                        & \multirow{7}{*}{20\%} & 89.23             & 3.15          & 76.39             & 1.42          & 87.04             & 2.15\\
        DTC                             &                       & 89.39             & 2.16          & 76.27             & 2.20          & 86.28             & 2.11\\
        CVRL*                           &                       & 90.15             & 2.01          & 77.33             & 2.18          & 88.12             & 2.41 \\
        MC-NET                          &                       & 90.11             & 2.02          & 78.17             & \textbf{1.55} & 87.83             & 1.52\\ 
        MC-NET+                         &                       & 91.09             & 1.71          & 79.37             & 1.72          & 88.51             & 1.54\\
        RCPS*                           &                       & 91.15             & 1.95          & 80.52             & 2.19          & 88.92             & 1.68\\
        \textbf{FBA-Net}                &                       & \textbf{91.31}    & \textbf{1.52} & \textbf{80.97}    & 1.59          & \textbf{89.81}    & \textbf{1.11}     \\ 
        \bottomrule
    \end{tabular}
\end{table}
\section{Experiments and Results}
\subsection{Dataset}
\textbf{(1) LA Dataset}\footnote[1]{\href{https://www.cardiacatlas.org/atriaseg2018-challenge/}{https://www.cardiacatlas.org/atriaseg2018-challenge/}} is the benchmark dataset from the 2018 MICCAI Atria Segmentation Challenge \cite{xiong2021global}. This dataset consists of 100 3D GE CMR images, including segmentation labels for the left atrium. The scans were acquired at an isotropic resolution of $0.625\times0.625\times0.625mm^3$. The dataset was split into two sets: 80 scans for training and 20 scans for evaluation. \textbf{(2) Pancreas-CT}\footnote[2]{\href{https://wiki.cancerimagingarchive.net/display/Public/Pancreas-CT}{https://wiki.cancerimagingarchive.net/display/Public/Pancreas-CT}} \cite{clark2013cancer} contains 82 patients of 3D abdominal contrast-enhanced CT scans. We partitioned this dataset into two sets: 62 samples for training and 20 samples for evaluation. \textbf{(3) ACDC dataset}\footnote[3]{\href{https://www.creatis.insa-lyon.fr/Challenge/acdc/databases.html}{https://www.creatis.insa-lyon.fr/Challenge/acdc/databases.html}} \cite{bernard2018deep} is a collection of cardiac MRIs, containing 100 short-axis cine-MRIs and three classes: left and right ventricle, and myocardium. Following \cite{wu2022mutual}, we applied a fixed data split, where 70, 10, and 20 patients data for training, validation, and testing sets, respectively. 
\subsection{Implementation}
All methods are implemented in PyTorch 1.12 with an NVIDIA 3090Ti GPU. To maintain consistency with the experiment setting outlined in \cite{you2022momentum}, we employed V-Net/U-Net as the backbone network and trained the models for 15,000 iterations. During the training process, we utilized SGD optimizer with a momentum value of 0.9 and weight decay of 0.005. The initial learning rate was set to 0.01. We set the batch size to 4, which contained 2 labeled scans and 2 unlabeled scans. Following previous works \cite{yu2019uncertainty}, we used the Dice similarity coefficient (DSC) and Average Surface Distance (ASD) to evaluate the segmentation performance. 
\begin{table}[t]
    \begin{minipage}{.5\linewidth}
        \centering
    
        \caption{The effectiveness of contrastive module as a plug-and-play. CM denotes the contrastive module.}
        \label{tab:tb2}
        \begin{tabular}{@{}lll@{}}
            \toprule
            Methods & Dice $\uparrow$ & ASD $\downarrow$ \\
            \midrule
            SSASSNet & 89.23 & 3.15 \\ 
            +CM      & 90.11(\textcolor{red}{+0.88}) & 2.52(\textcolor{red}{-0.63}) \\
            \midrule
            DTC      & 89.39 & 2.16 \\
            +CM      & 90.40(\textcolor{red}{+1.01}) & 2.12(\textcolor{red}{-0.04}) \\
            \midrule
            MC-NET   & 90.11 & 2.02 \\
            +CM      & 90.47(\textcolor{red}{+0.36}) & 2.06(\textcolor{blue}{+0.04}) \\
            \bottomrule
        \end{tabular}
    \end{minipage}%
    \begin{minipage}{.45\linewidth}
        \centering
        \caption{Comparison between InfoNCE Loss and FBA Loss.}
        \label{tab:tb3}
        \begin{tabular}{@{}lll@{}}
            \toprule
            Methods         & Dice $\uparrow$ & ASD $\downarrow$ \\
            \midrule
            InfoNCE Loss     & 91.07 & 1.67 \\
            FBA Loss        & \textbf{91.31} & \textbf{1.52} \\
            \bottomrule
        \end{tabular}
    \end{minipage}
\end{table}

\begin{figure}[t]
\includegraphics[width=\textwidth]{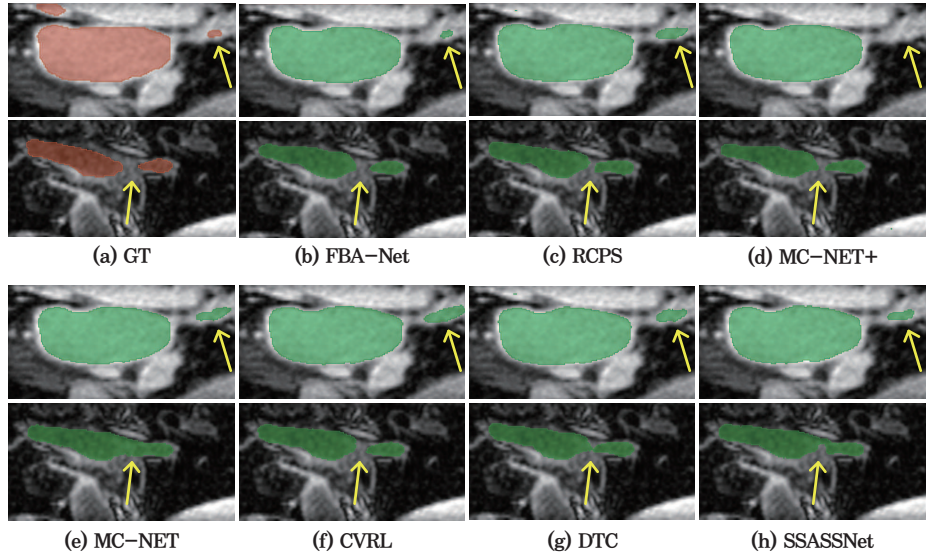}
\caption{Visual comparison between FBA-Net and the state-of-the-art methods trained with 20\% labeled data. The areas highlighted in red and green represent the ground truth and predicted regions, respectively.} \label{fig2}
\end{figure}
\subsection{Results}
\subsubsection{Quantitative result.} This section presents the quantitative results of our proposed approach for medical image segmentation on three datasets: LA, Pancreas-CT, and ACDC. As shown in Table \ref{tab:tb1}, our approach is capable of extracting distinguishing features using a minimal number of labels. This results in performance metrics that closely approximate those attained by the fully-supervised method, i.e., 91.31 vs 91.62 on the LA dataset. FBA-Net surpasses other contrastive learning methods, specifically CVRL and RCPS, across most metrics over all datasets. The results on the ACDC dataset outperform all state-of-the-art methods such as SSASSNet, DTC, CVRL, MC-NET, MC-NET+, and RCPS. This indicates the versatile applicability of our approach, which can be effectively employed in binary-class segmentation and extends to multi-class segmentation as well. Significantly, the ASD score yielded by our method is considerably lower than the previously lowest scores (0.71 vs 1.40 and 1.11 vs 1.52), demonstrating the method's efficacy. These findings highlight the usefulness of our approach in improving the segmentation accuracy of medical images through the integration of tailored contrastive learning.\\
\subsubsection{Qualitative result.} The visualizations in Fig~\ref{fig2} illustrate the segmentation results for FBA-Net and other methods. In particular, FBA-Net has produced highly accurate segmentation results, closely mirroring the ground truths and surpassing other approaches. The results indicate that FBA-Net can efficiently segment even the most challenging parts of the images, as pointed out by the yellow arrow. Notably, other methods either over-segmented or completely missed the regions indicated by the arrow in the first sample, while FBA-Net successfully segmented these areas. These results highlight the precision of the FBA-Net in medical image segmentation, especially in challenging scenarios.
\subsubsection{Ablation Study} In order to assess the utility of our proposed contrastive module as a plug-and-play component, we apply it to other semi-supervised methods including SSASSNet, DTC, and MC-Net. Table \ref{tab:tb2} reveals that the addition of our contrastive module leads to an improvement in segmentation performance. In particular, the Dice scores for the three models show respective increases of 0.88\%, 1.01\%, and 0.36\%. We further demonstrate the effectiveness of our tailored FBA loss in distinguishing representations for segmentation by comparing it with the InfoNCE loss on the LA dataset. The FBA loss shows superior performance across all metrics when compared to InfoNCE loss, a common choice in other contrastive learning methods such as CVRL and RCPS. This highlights not only the significance of differentiating feature representation but also the enhanced effectiveness of our loss function in medical image segmentation tasks.   
\section{Conclusion}
In this paper, we proposed a contrastive learning approach that focuses on learning the foreground and background features separately for accurate segmentation. By utilizing the contrastive module, our method has the potential to greatly reduce the costs and time associated with manual annotation, which could have a significant impact by enabling the rapid development of diagnostic tools and treatments. Additionally, our approach demonstrated state-of-the-art performance. The proposed approach can be extended to other medical image segmentation tasks where foreground-background separation is crucial for accurate segmentation.
%
%
%
\bibliographystyle{splncs04}
\bibliography{paper}
\end{document}